\documentclass[10pt,twocolumn,letterpaper]{article}

\usepackage{cvpr}
\usepackage{times}
\usepackage{epsfig}
\usepackage{graphicx}
\usepackage{amsmath}
\usepackage{amssymb}
\usepackage{bbm}
\usepackage{multirow}

\usepackage[pagebackref=true,breaklinks=true,letterpaper=true,colorlinks,bookmarks=false]{hyperref}

\cvprfinalcopy 

\def\cvprPaperID{423} 

\ifcvprfinal\fi
\begin{document}

\title{Perceive Where to Focus: Learning Visibility-aware Part-level Features\\ for Partial Person Re-identification}

\author{%
Yifan Sun$^1$, Qin Xu$^1$, Yali Li$^1$, Chi Zhang$^2$, Yikang Li$^1$, Shengjin Wang$^1$\thanks{Corresponding author.},  Jian Sun$^2$\\
{$^1$Tsinghua University}          
{$^2$Megvii Technology}\\
{\texttt{\small{\{sunyf15, xuq16, liyk11\}@mails.tsinghua.edu.cn}} \hspace{0.5cm}}
{\texttt{\small{\{liyali13, wgsgj\}@tsinghua.edu.cn}}\hspace{0.5cm}}\\
{\texttt{\small{\{zhangchi, sunjian\}@megvii.com}}}
}

\maketitle

\begin{abstract}

This paper considers a realistic problem in person re-identification (re-ID) task, \emph{i.e.}, partial re-ID. Under partial re-ID scenario, the images may contain a partial observation of a pedestrian. 
If we directly compare a partial pedestrian image with a holistic one, the extreme spatial misalignment significantly compromises the discriminative ability of the learned representation. We propose a Visibility-aware Part Model (VPM), which learns to perceive the visibility of regions through self-supervision. The visibility awareness allows VPM to extract region-level features and compare two images with focus on their shared regions (which are visible on both images). 
VPM gains two-fold benefit toward higher accuracy for partial re-ID. On the one hand, compared with learning a global feature, VPM learns region-level features and benefits from fine-grained information. On the other hand, with visibility awareness, VPM is capable to estimate the shared regions between two images and thus suppresses the spatial misalignment. Experimental results confirm that our method significantly improves the learned representation and the achieved accuracy is on par with the state of the art. 
 
\end{abstract}

\section{Introduction}
Person re-identification (re-ID) aims to spot the appearances of same person in different observations by measuring the similarity between the query image and the gallery images (\emph{i.e.}, the database). In spite that the re-ID research community has achieved significant progress during the past few years, re-ID systems are still faced with a series of realistic difficulties. A prominent challenge is the partial re-ID problem \cite {DBLP:conf/iccv/ZhengLXLLG15, HeLX2018Partial, DBLP:conf/cvpr/ZhengGX11},  which requires accurate retrieval with partial observation of the pedestrian. More concretely, in realistic re-ID systems, a pedestrian may happen to be partially occluded or be walking out of the field of camera view, and the camera fails to capture the holistic pedestrian. 

\begin{figure}[t]
\setlength{\abovecaptionskip}{-1.2cm} 
\setlength{\belowcaptionskip}{-0.4cm}
\begin{center}
\includegraphics[width=0.8\linewidth]{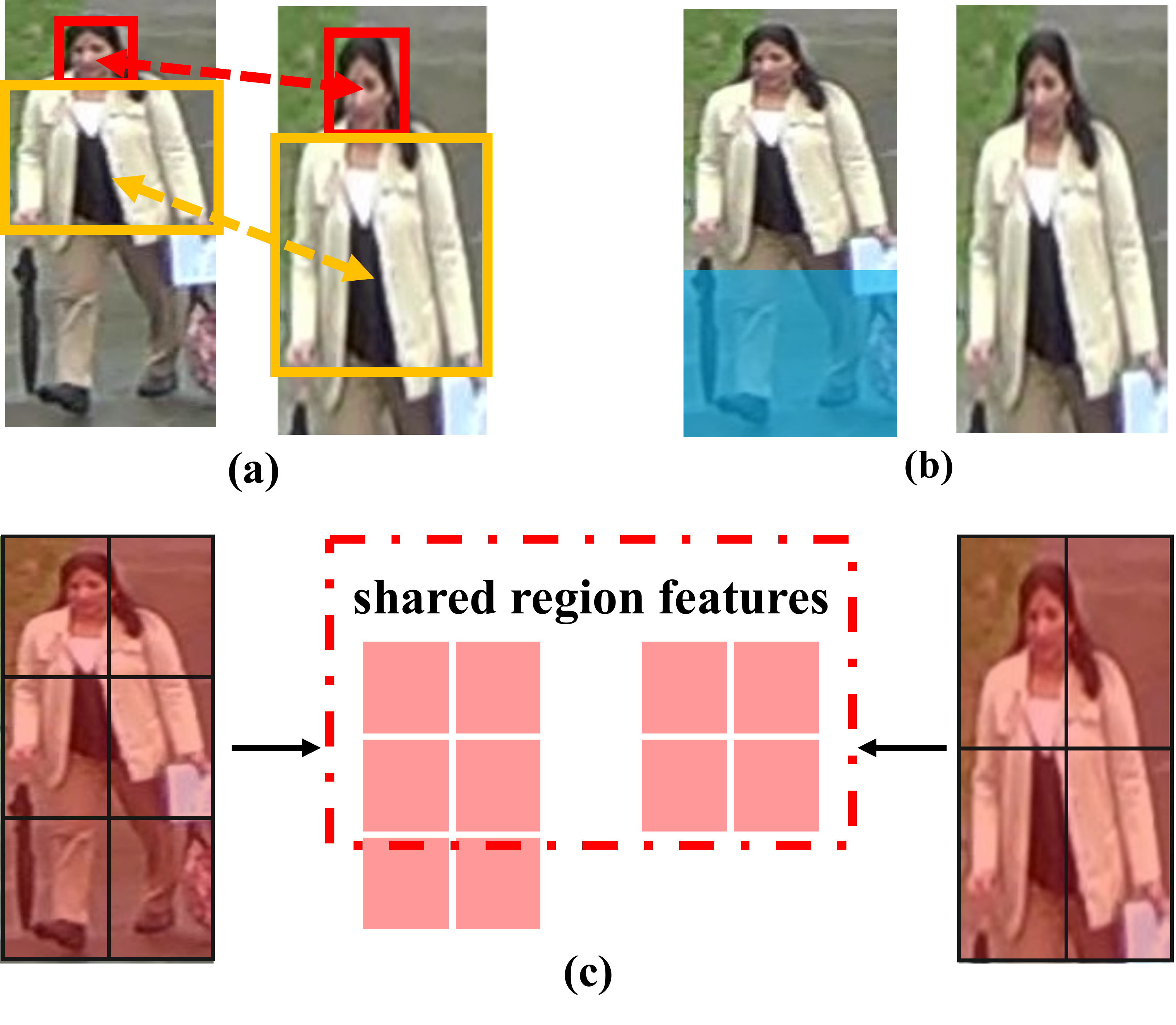}
\end{center}
   \caption{Two challenges related to partial-re-ID and our solution with the proposed VPM. (a) aggravation of spatial misalignment, (b) distracting noises from unshared regions (the blue region on the left image) and (c) VPM locates visible regions on a given image and extracts region-level features. With visibility-awareness, VPM compares two images by focusing on their shared regions.}
\label{fig:misalign}
\end{figure}

Intuitively, partial re-ID increases the difficulty to make correct retrieval. Analytically,  we find that partial re-ID raises two more unique challenges, compared with the holistic person re-ID, as illustrated in Fig. \ref{fig:misalign}. 
\begin {itemize}
\item First, partial re-ID aggravates the spatial misalignment between probe and gallery images. Under holistic re-ID setting, the spatial misalignment mainly originates from the articulate movement of pedestrian and the viewpoint variation. Under partial re-ID setting, even when two pedestrian with same pose are captured from same viewpoints, there still exists severe spatial misalignment between the two images (Fig. \ref{fig:misalign} (a)).
\item Second, when we directly compare a partial pedestrian against a holistic one, the unshared body regions in the holistic pedestrian become distracting noises, rather than discriminative clues. We note that the same situation also happens when any two compared images contain different proportion of the holistic pedestrian (Fig. \ref{fig:misalign} (b)). 
\end {itemize}

We propose the Visibility-aware Part Model (VPM) for partial re-ID. VPM avoids or alleviates the two unique difficulties related to partial re-ID by focusing on their shared regions, as shown in Fig. \ref{fig:misalign} (c). More specifically, we first define a set of regions on the holistic person image. During training, given partial pedestrian images, VPM learns to locate all the pre-defined regions on convolutional feature maps. After locating each region, VPM perceives which regions are visible and learns region-level features. During testing, given two images to be compared, VPM first calculates the local distances between their shared regions and then concludes the overall distance.   

VPM gains two-fold benefit toward higher accuracy for partial re-ID. On the one hand, compared with learning a global feature, VPM learns region-level features and thus benefits from fine-grained information, which is similar to the situation in holistic person re-ID \cite{PCB_ECCV, cvpr2018:SPReID}. On the other hand, with visibility-awareness, VPM is capable to estimate the shared regions between two images and thus suppresses the spatial misalignment as well as the noises originated from unshared regions. Experimental results confirm that VPM achieves significant improvement on partial re-ID accuracy, compared with a global feature learning baseline \cite{DBLP:journals/corr/ZhengYH16}, as well as a strong part-based convolutional baseline \cite{PCB_ECCV}. The achieved performance are on par with the state of the art. 

Moreover, VPM is featured for employing self-supervision for learning the region visibility awareness. We randomly crop partial pedestrian images from the holistic ones and automatically generate region labels, yielding the so-called self-supervision. Self-supervision enables VPM to learn locating pre-defined regions. It also helps VPM to focus on visible regions during feature learning, which is critical to the discriminative ability of the learned features, as to be accessed in Section \ref{sec:importance}.


The main contributions of this paper are summarized as follows:
\begin {itemize}
\item We propose a visibility-aware part model (VPM) for partial re-ID task. VPM learns to locate the visible regions on pedestrian images through self-supervision. Given two images to be compared, VPM conducts a region-to-region comparison within their shared regions, and thus significantly suppresses the spatial misalignment as well as the distracting noises originated from unshared regions. 
\item We conduct extensive partial re-ID experiments on both synthetic datasets and realistic datasets and validate the effectiveness of VPM. On two realistic dataset, Partial-iLIDs and Partial-ReID, VPM achieves performance on par with the state of the art. So far as we know, few previous works on partial re-ID reported the performance on synthetic large-scale datasets \emph{e.g.}, Market-1501 or DukeMTMC-ReID. We experimentally validate that VPM can be easily scaled up to large-scale (synthetic) partial re-ID datasets, due to its fast matching capacity. 
\end {itemize}

\section {Related Works}
\subsection {Deeply-learned part features for re-ID}
Deep learning methods currently dominate the re-ID research community with significant superiority on retrieval accuracy \cite{DBLP:journals/corr/ZhengYH16}. Recent works \cite{PCB_ECCV, cvpr2018:SPReID, Wei2017GLAD, Zhao2017Deeply,Su2017Pose,Yao2017Deep,Liu2017HydraPlus} further advance the state of the art on holistic person re-ID, through learning part-level deep features. For example, Wei \emph{et al.} \cite{Wei2017GLAD}, Kalayeh \emph{et al.} \cite{cvpr2018:SPReID} and Sun \emph {et al.} \cite{PCB_ECCV} extract several region parts, with pose estimation \cite{pose:Long2015Fully,pose:CPM,pose:DeeperCut,pose:hourglass,pose:Cao2016Realtime}, human parsing \cite{parsing:Chen2018DeepLab,parsing:self_supervised_human_parsing} and uniform partitioning, respectively. Then they learn a respective feature for each part and assemble the part-level features to form the final descriptor. 
These progresses motivate us to extend learning part-level features to the specified problem of partial re-ID. 

However, learning part-level features does \emph {not} naturally improve partial re-ID. We find that PCB \cite{PCB_ECCV}, which maintains the latest state of the art on holistic person re-ID, encounters a substantial performance decrease when applied in partial re-ID scenario. The achieved retrieval accuracy even drops below the global feature learning baseline (to be accessed in Sec. \ref{sec:largedata}). Arguably, it is because part models rely on precisely locating each part and are inherently more sensitive to the severe spatial misalignment problem in partial re-ID. 

Our method is similar to PCB in that both methods perform uniform division instead of semantic body parts for part extraction. Moreover, similar to SPReID \cite{cvpr2018:SPReID}, our method also uses probability maps to extract each part during inference. However, while SPReID requires an extra human parser and human parsing dataset (strong supervision) for learning part extraction, our method relies on self-supervision.  During matching stage, both PCB and SPReID adopt the common strategy of concatenating part features. In contrast, VPM first measures the region-to-region distance and then conclude the overall distance by dynamically crediting the local distances with high visibility confidence.

\begin{figure*}[t]
\setlength{\abovecaptionskip}{-0.2cm} 
\setlength{\belowcaptionskip}{-0.1cm}
\begin{center}
\includegraphics[width=0.95\linewidth]{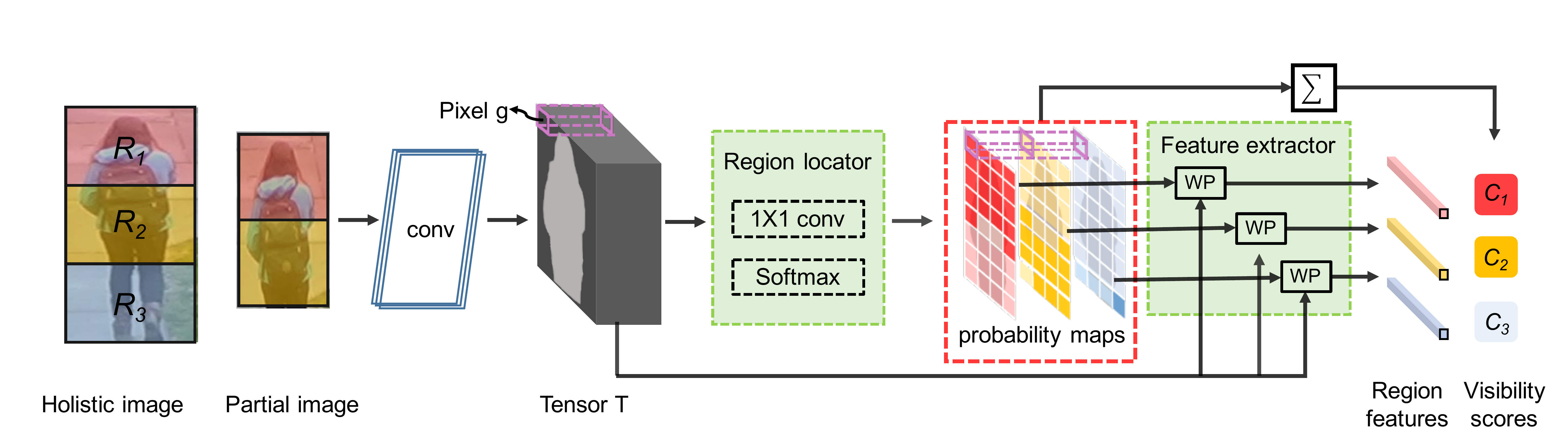}
\end{center}
   \caption{The structure of VPM. We first define $p=m\times n$ ($3\times 1$ in the figure for instance) densely aligned rectangle regions on the holistic pedestrian. VPM resizes a partial pedestrian image to fixed size, inputs it into a stack of convolutional layers (``conv'') and transforms it into a 3D tensor $T$. Upon $T$, VPM appends a region locator to discover each regions through pixel-wise classification. By predicting a probability of belonging to each region for every pixel $g$, the region locator generates $p$ probability maps to infer the location of each region. It also generates $p$ visibility scores through ``$\sum$'' operation over each probability map. Given the predicted probability maps, the feature extractor extracts a respective feature for each pre-defined region through weighted pooling (``WP''). VPM, as a whole, outputs $p$ region-level features and $p$ visibility scores for inference.}
\label{fig:structure}
\end{figure*}

\subsection {Self-supervised learning}
Self-supervision learning is a specified unsupervised learning approach. It explores the visual information to automatically generate surrogate supervision signal for feature learning \cite{Noroozi2016Self, DBLP:conf/iccv/0004HG17,Larsson2016Color, Doersch2015Self, DBLP:conf/icml/LeRMDCCDN12}. Larsson \emph {et al.} \cite{Larsson2016Color} train the deep model to predict per-pixel color histograms and consequentially facilitate automatic colorization. Doersch \emph{et al.} \cite{Doersch2015Self} and Noroozi \emph{et al.} \cite{Noroozi2016Self} propose to predict the relative position of image patches. Gidaris \emph{et al.} train the deep model to recognize the rotation applied to original images. 

Self-supervision is an elemental tool in our work. We employ self-supervision to learn visibility awareness. VPM is especially close to \cite{Doersch2015Self} and \cite{Noroozi2016Self} in that all the three methods employ the position information of patches for self-supervision. However, VPM significantly differs from them in the following aspects.

\textbf{Self-supervision signal.} 
\cite{Doersch2015Self} randomly samples a patch and one of its eight possible neighbors, and then trains the deep model to recognize the spatial configuration. Similarly, \cite{Noroozi2016Self} encodes the neighborhood relationship into a jigsaw puzzle. Different from \cite{Doersch2015Self} and \cite{Noroozi2016Self}, VPM does not explore the spatial relationship between multiple images or patches. 
VPM pre-defines a division on the holistic pedestrian image and then assigns an independent label to each region. Then VPM learns to directly predict which regions are visible on a partial pedestrian image, without comparing it against the holistic one. 

\textbf{Usage of the self-supervision.} Both \cite{Doersch2015Self} and \cite{Noroozi2016Self} transfer the model trained through self-supervision to the object detection or classification task. In comparison, VPM utilizes self-supervision in a more explicit manner: with the visibility awareness gained from self-supervision, VPM decides which regions to focus when comparing two images.

\section {Proposed Method}
\subsection{Structure of VPM}\label{sec:structure}

VPM is designed as a fully convolutional network, as illustrated in Fig. \ref{fig:structure}. It takes a pedestrian image as the input and outputs a constant number of region-level features, as well as a set of visibility scores indicating which regions are visible on the input image. 

We first define $p=m\times n$ densely aligned rectangle regions on the holistic pedestrian image through uniform division. Given a partial pedestrian image, we resize it to a fixed size, \emph{i.e.}, $H\times W$ and input it into VPM. Through a stack of convolutional layers (``conv'' in Fig. \ref{fig:structure}, which uses all the convolutional layers in ResNet-50 \cite{DBLP:conf/cvpr/HeZRS16}), VPM transfers the input image into a 3D tensor $T$. The size of $T$ is $c \times h \times w$ (which are the number of channels, height and width, respectively), and we view the $c-dim$ vector $g$ as a pixel on $T$. On tensor $T$, VPM appends a region locator and a region feature extractor. The region locator discovers regions on tensor $T$. Then the region feature extractor generates a respective feature for each region.

\textbf{A region locator} perceives which regions are visible and predicts their locations on tensor $T$. To this end, the region locator employs a $1\times1$ convolutional layer and a following Softmax function to classify each pixel $g$ on $T$ into the pre-defined regions, which in formulated by, 
\begin{equation}\label{eq:region}
P(R_i|g) = softmax(W^Tg) = \frac{\exp{W_i^Tg}}{\sum\limits_{j=1}^p{\exp{W_j^Tg}}},
\end{equation}
where $P(R_i|g)$ is the predicted probability of $g$ belonging to $R_i$, $W$ is the learnable weight matrix of the $1\times1$ convolutional layer, $p$ is the total number of pre-defined regions.

By sliding over every pixel $g$ on $T$, the region locator predicts $g$ as belonging to all the pre-defined regions with corresponding probabilities, and thus gets $p$ probability maps (one $h\times w$ map for each region), as shown in Fig. \ref{fig:structure}. Each probability map indicates the location of a corresponding region on $T$, which allows region feature extraction.

The region locator also predicts the visibility score $C$ for each region, by accumulating $P(R_i|g)$ over all the $g$ on $T$, which is formulated by, 
 \begin{equation}\label{eq:visibility}
 C_i = \sum\limits_{f \in T}{P(R_i|g)},
\end{equation}
 
 Eq. \ref{eq:visibility} is natural in that if considerable pixels on $T$ belongs to $R_i$ (with large probability), it indicates that $R_i$ is visible on the input image and is assigned with a relatively large $C_i$. In contrast, if a region is actually invisible, the region locator will still return a probability map (with all the values approximating 0). In this case, $C_i$ will be very small, indicating possibly-invisible region. The visibility score is important for calculating the distance between two images, as to be detailed in Section \ref{sec:testing}.
 
\textbf{A region feature extractor} generates a respective feature $f$ for a region by weighted pooling, which is formulated by,
 \begin{equation}\label{eq:feature}
 f_i = \frac{\sum\limits_{g \in T}{P(R_i|g)g}}{C_i}, \forall i \in \{1,2,\cdots,p\},
 \end{equation}
 where the division of $C_i$ is to maintain the norm invariance against the size of the region. 
 
The region locator returns a probability map for each region, even if the region is actually invisible on the input image. Correspondingly, we can see from Eq. \ref{eq:feature} that the region feature extractor always generates a constant number (\emph{i.e.}, $p$) of region features for any input image.
 
 \subsection{Employing VPM}\label{sec:testing}
Given two images to be compared, \emph{i.e.}, $I^k$ and $I^l$, VPM extracts their region features and predicts the region visibility scores through Eq. \ref{eq:feature} and Eq. \ref{eq:visibility}, respectively.
With region features and region visibility scores $\{f_i^{k}, C_i^k\}$ , $\{f_i^{l}, C_i^l\}$, VPM first calculates region-to-region euclidean distances $D_i^{kl} = \Vert f_i^k - f_i^l\Vert (i = 1, 2, \cdots, p)$. Then VPM concludes the overall distance from the local distances by, 
\begin{equation}\label{eq:distance} 
D^{kl} = \frac{\sum\limits_{i=1}^p{C_i^kC_i^lD_i^{kl}}}{\sum\limits_{i=1}^p{C_i^kC_i^l}}. 
\end{equation}

In Eq. \ref{eq:distance}, the visible regions are with relative large visibility scores. The local distances between shared regions are highly credited by VPM and thus dominate the overall distance $D^{kl}$. In contrast, if a region is invisible in any one of the compared images, its region feature is considered as unreliable and the corresponding local distance contributes little to $D^{kl}$. 

Employing VPM adds very light computational cost, compared with popular part-based deep learning methods \cite{PCB_ECCV, Zhao2017Deeply,cvpr2018:SPReID}. While some prior partial re-ID methods require pairwise comparison before feature extraction and may have efficiency problems, VPM presents high scalability, which allows experiments on large re-ID datasets such as Market-1501 \cite{DBLP:conf/iccv/ZhengSTWWT15} and DukeMTMC-reID \cite{zheng2017unlabeled}, as to be accessed in Section \ref{sec:largedata}.

\subsection{Training VPM}
Training VPM consists of training the region classifier and training the region feature extractor. The region classifier and the region feature extractor share the convolutional layers before tensor $T$, and are trained end to end in a multi-task training manner. Training VPM is also featured for employing auxiliary self-supervision.

\textbf{Self-supervision} is critical to VPM. It supervises VPM to learn region visibility awareness, as well as to focus on visible regions during feature learning. Specifically, given a holistic pedestrian image, we randomly crop a patch and resize it to $H \times W$. The random crop operation excludes several pre-defined regions and the remaining regions are reshaped during the resizing. Then, we project the regions on the input image to tensor $T$ through ROI projection \cite{kaiming14ECCV, Ren2015Faster}. To be concrete, let us assume a region with its up-left corner located at $(x_1, y_1)$ and its bottom-right corner located at $(x_2, y_2)$ on the input image. Then the ROI projection defines a corresponding region on tensor $T$ with its up-left corner located at $(\left[x_1/S\right], \left[y_1/S\right])$ and its right-bottom corner located at $(\left[x_2/S\right], \left[y_2/S\right])$, in which the $\left[\bullet\right]$ denotes the rounding and $S$ is the down-sampling rate from input image to $T$. Finally, we assign every pixel $g$ on $T$ with a region label $L (L \in {1, 2, \cdots, p})$ to indicate which region $g$ belongs to. We also record all the visible regions in a set $V$. As we will see, self-supervision contributes to training VPM in the following three aspects:

\begin{itemize}
\item First, self-supervision generates the ground truth of region labels for training the region locator.
\item Second, self-supervision enables VPM to focus on visible regions when learning feature through classification loss (cross-entropy loss). 
\item Finally, self-supervision enables VPM to focus on the shared regions when learning features through triplet loss.
\end{itemize}

Without the auxiliary self-supervision, VPM encounters dramatic performance decrease, as to be accessed in Section \ref{sec:importance}.

 \textbf{The region locator} is trained through cross-entropy loss with the self-supervision signal $L$ as the ground truth, which is formulated by,
\begin{equation}
L_{R} = - \sum\limits_{g \in T}{\mathbbm{1}_{i=L}log(P(R_i|g))},
\end{equation}
where $\mathbbm{1}_{R_i=L}$ returns 1 only when $i$ equals the ground truth region label $L$ and returns 0 in any other cases. 

\begin{figure}[t]
\setlength{\abovecaptionskip}{-1.2cm} 
\setlength{\belowcaptionskip}{-0.4cm}
\begin{center}
\includegraphics[width=.9\linewidth]{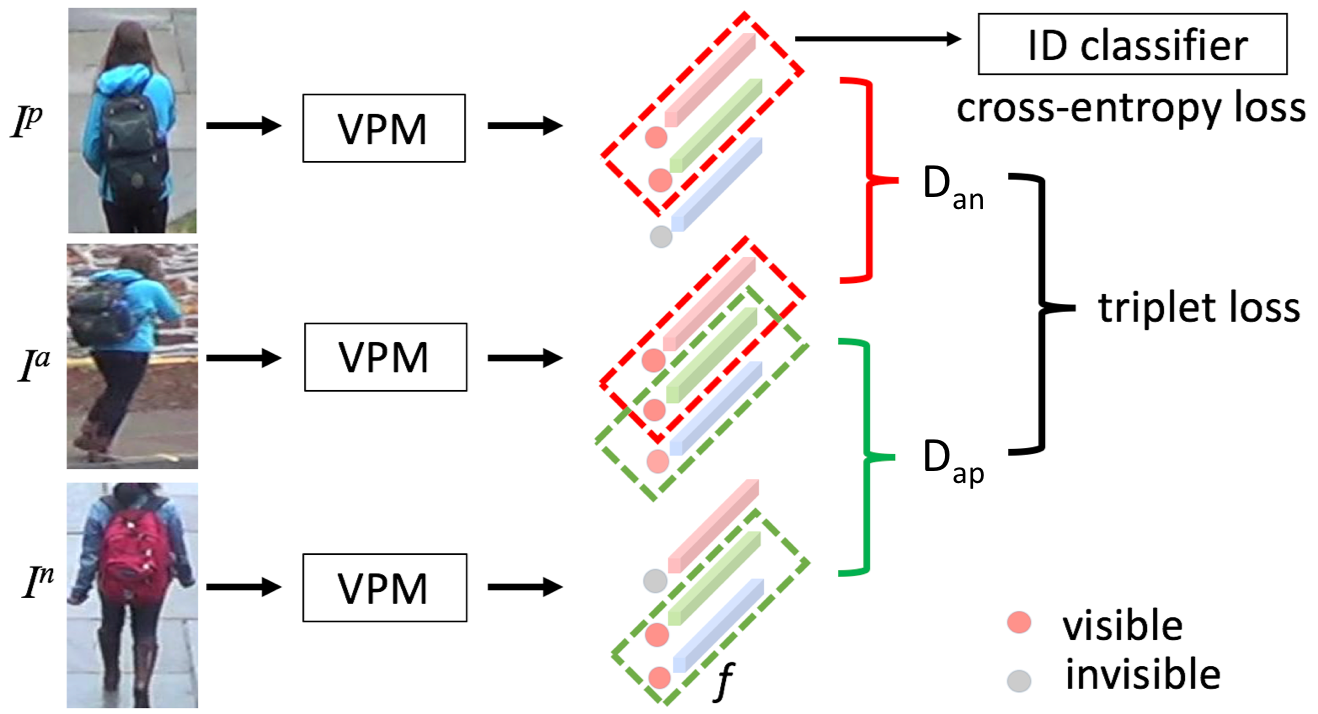}
\end{center}
   \caption{VPM learns region-level features with auxiliary self-supervision. Only features corresponding to visible regions contribute to the cross-entropy loss. Only features corresponding to shared regions contribute to the deducing of triplet loss.}
\label{fig:loss}
\end{figure}

\textbf{The region feature extractor} is trained with the combination of cross-entropy loss and triplet loss, as illustrated in \ref{fig:loss}.
 Recall that the region feature extractor always generates $p$ region features for any input image. It leads to a nontrivial problem during feature learning: only features of visible regions should be allowed to contribute to the training losses. With self-supervision signals $V$, we dynamically select the visible regions for feature learning.   

\emph{The cross-entropy loss} is commonly used in learning features for pedestrian under the IDE \cite{zheng2016mars} mode. We append a respective identity classifier \emph{i.e.}, $IP_i(f_i) (i = 1, 2, \cdots, p)$ upon each region feature $f_i$, to predict the identity of training images. The identity classifier consists of two sequential fully-connected layers and a Softmax function. The first fully-connected layers reduces the dimension of the input region feature, and the second one transforms the feature dimension to $K$ ($K$ is the total identities of training images).  Then the cross-entropy loss is formulated by,
\begin{equation}\label{eq:IDE}
L_{ID} = - \sum\limits_{i \in V}{\mathbbm{1}_{k=y}log(softmax(IP_i(f_i)))},
\end{equation}
where $k$ is the predicted identity and $y$ is the ground truth label. With Eq. \ref{eq:IDE}, self-supervision enforces focus on visible regions for learning region features through cross-entropy loss. 


\emph{The triplet loss} pushes the features from a same pedestrian close to each other and pulls the features from different pedestrians far away. 
Given a triplet of images, \emph{i.e.}, an anchor image $I^a$, a positive image $I^p$ and a negative image $I^n$, we define a region-selective triplet loss derived from the canonical one by, 

\begin{equation}\label{eq:triplet}
\begin{aligned}
L_{tri} &= \left[D^{ap}
- D^{an}
+ \alpha\right]_+,  \\
D^{ap} &= \frac{\sum\limits_{i \in (V^a \cap V^p)}{\lVert{f_i^a-f_i^p}\rVert}}{\lvert{V^a \cap V^p}\rvert},\\
D^{an} &= \frac{\sum\limits_{i \in (V^a \cap V^n)}{\lVert{f_i^a-f_i^n}\rVert}}{\lvert{V^a \cap V^n}\rvert},\\
\end{aligned}
\end{equation}
where $f_i^a$, $f_i^p$ and $f_i^n$ are the region features from anchor image, positive image and negative image, respectively. $V^a$, $V^p$ and $V^n$ are the visible region sets for anchor image, positive image and negative image, respectively. $\lvert{\bullet}\rvert$ denotes the operation of counting the elements of a set, \emph{i.e.}, the number of shared regions in the two compared images. $\alpha$ is the margin for training triplet, and is set to $1$ in our implementation.

With Eq. \ref{eq:triplet}, self-supervision enforces focus on the shared regions when calculating the distances of two images. 

The overall training loss is the sum of region prediction loss, the identity classification loss and the region-selective triplet loss, which is formulated by,
\begin{equation}
L = L_R + L_{ID} + L_{tri}
\end{equation}

We also note that Eq. \ref{eq:distance} and Eq. \ref{eq:triplet} share a similar pattern. Training with the modified triplet loss (Eq. \ref{eq:triplet}) mimics the matching strategy (Eq. \ref{eq:distance}) and is thus specially beneficial (to be detailed in Table \ref{tab:ablation}). The difference is that, during training, the focus is enforced through ``hard'' visibility labels, while during testing, the focus is regularized through predicted ``soft'' visibility scores. 

\setlength{\tabcolsep}{6.5pt}
\begin{table*}[t]
\begin{center}
\begin{tabular}{l|c|ccc c|ccc c|ccc c}
\hline
\multicolumn{1}{l|}{\multirow{2}{*}{Dataset }}&\multicolumn{1}{c|}{\multirow{2}{*}{$\gamma$}}&\multicolumn{4}{c|}{baseline} & \multicolumn{4}{c|}{PCB} & \multicolumn{4}{c}{VPM}\\ 
\cline{3-14}
\multicolumn{1}{c|}{}&\multicolumn{1}{c|}{}&\multicolumn{1}{c}{R-1}&R-5&R-10&{mAP}&{R-1}&R-5&R-10&{mAP}&{R-1}&R-5&R-10&{mAP}\\
\hline

            & 0.5 &64.5&82.2&88.1&44.4  &0.9 &3.2 &5.6 &1.7  &70.9&86.5&92.1&48.8\\
            & 0.6 &79.0&91.4&94.3&57.9  &8.1 &16.5 &23.2 &6.6  &84.4&94.3&96.1&62.5\\
Market-1501 & 0.7 &83.9&93.9&95.9&63.7  &36.5 &58.9 &67.4 &26.8  &88.2&95.8&97.2&71.7\\
            & 0.8 &85.7&94.3&96.4&66.1  &71.9 &87.3 &91.4 &56.8  &90.1&95.8&97.7&74.7\\
            & 0.9 &87.1&95.5&97.4&67.8  &88.8 &95.8 &97.1 &77.2  &91.7&96.6&98.0&78.7\\
            & 1.0 &86.8&95.3&97.4&67.7  &93.4 &97.8 &98.4 &83.0    &93.0&97.8&98.8&80.8\\
\hline
              &0.5  &65.0&81.1&86.7&47.2  &5.0 &10.1 &13.6 & 4.0     &69.5&83.1&87.9&52.2\\
              &0.6  &76.2&87.3&90.4&55.4  &13.1 &25.6 &33.5 &10.5    &78.2&89.0&91.3&60.9\\
DukeMTMC-reID &0.7  &76.3&87.3&90.6&90.6  &35.9 &57.0 &65.4 &28.4    &80.3&89.5&92.0&63.1\\
              &0.8  &76.3&88.3&91.9&58.8  &64.0 &82.6 &87.7 &52.3    &80.3&89.3&92.4&63.5\\
              &0.9  &77.0&88.1&91.7&59.0 &81.6 &90.4 &93.0 &70.3    &81.7&90.9&93.1&70.7\\
              &1.0  &76.2&87.3&91.2&58.6  &84.1 &92.4 &94.5 &73.2    &83.6&91.7&94.2&72.6\\
  
\hline

\end{tabular}
\caption{Comparison between VPM, baseline and PCB. For VPM, we use $p=6\times 1$ pre-defined regions. For PCB, we adopt the code released by the authors and append an extra triplet loss, for fair comparison with VPM. On Market-1501, the extra triplet loss enables PCB to gain +5.6\% mAP over the original 77.4\% reported by the authors \cite{PCB_ECCV}. }
\label{tab:large}
\end{center}
\end{table*}

\section{Experiment}
\subsection{Settings}
\textbf{Datasets}. We use four datasets, \emph{i.e.}, Market-1501 \cite{DBLP:conf/iccv/ZhengSTWWT15}, DukeMTMC-reID \cite{ristani2016MTMC,zheng2017unlabeled}, Partial-REID and Partial-iLIDS, to evaluate our method. Market-1501 and DukeMTMC-reID are two large scale holistic re-ID dataset. The Market-1501 dataset contains 1,501 identities observed from 6 camera viewpoints, 19,732 gallery images and 12,936 training images detected by DPM \cite{felzenszwalb2008discriminatively}. The DukeMTMC-reID dataset contains 1,404 identities, 16,522 training images, 2,228 queries, and 17,661 gallery images. We crop certain patches from the query images during testing stage to imitate the partial re-ID scenario and get a comprehensive evaluation of our method on large-scale (synthetic)  partial re-ID datasets. We note that few prior works on partial re-ID evaluated their methods on large-scale dataset, mainly because of low computing efficiency. Partial-REID \cite{DBLP:conf/iccv/ZhengLXLLG15} and Partial-iLIDS \cite{DBLP:conf/cvpr/ZhengGX11} are two commonly-used datasets for partial re-ID. Partial-REID contains 600 images and 60 identities, every one of which has 5 holistic images and 5 partial images. Partial-iLIDS is derived from iLIDS \cite{DBLP:conf/eccv/WangGZW14}, which is collected in an airport and the lower-body of a pedestrian is frequently occluded by the luggage. Partial-iLIDS crops the non-occluded region from these images and get 238 images from 119 identities. Both Partial-REID and Partial-iLIDS offer only testing images. When evaluating our method on these two public datasets, we train VPM on the training set of Market-1501, for fair comparison with other competitive methods, including MTRC \cite{DBLP:journals/pami/LiaoJL13}, AMC+SWM \cite{DBLP:conf/iccv/ZhengLXLLG15}, DSR \cite{HeLX2018Partial}, and SFR \cite{DBLP:journals/corr/abs-1810-07399}.

\textbf{Implementation Details}. 
Training VPM relies on the assumption that the original training images all contain holistic pedestrian and the pedestrian are tightly bounded by bounding boxes. On two holistic re-ID datasets, Market-1501 and DukeMTMC-reID, there do exist some images which contain either partial pedestrian or oversized bounding boxes. We consider these images as tolerable noises. 

To generate the partial image for training VPM, we crop a patch from the holistic image with random area ratio $\gamma$. We set $\gamma$ to be uniformly distributed between $0.5$ and $1$. VPM is not necessarily bounded with any specified crop strategy and we may consider the prior knowledge for optimization. 
We argue that choosing the detailed crop strategy according to the realistic condition is reasonable because partial re-ID is a realistic challenge, and the occlusion fashion is usually predictable. We also experimentally validate that choosing an appropriate crop strategy to imitate the confronted partial re-ID condition benefits the retrieval accuracy, as to be detailed in Section \ref{sec:sota}. That being said, VPM is still general in that it may adopt any crop strategy to conduct self-supervision. 

VPM is trained with the combination of cross-entropy loss and triplet loss. We pre-train the model for 50 epochs with single cross-entropy loss because it helps VPM to converge faster and better. Then we append the triplet loss and fine-tune the model for another 80 epochs. In both the pre-training and the fine-tuning stages, we use standard Stochastic Gradient Descent (SGD) optimization strategy, initialize the learning rate as 0.1 and decay it to 0.01 after 30 epochs. During the fine-tuning stage, we construct each mini-batch with 64 images from 8 identities (8 images per identity) and use the hard mining strategy \cite{Hermans2017DefenseTriplet} for deducing the triplet loss. 


\subsection{Evaluation on large-scale partial re-ID datasets}\label{sec:largedata}
We evaluate the effectiveness of VPM with experiment on the synthetic partial datasets derived from two large-scale re-ID datasets, Market-1501 and DukeMTMC-reID. We differ the ratio $\gamma$ of the cropped patches from 0.5 to 1.0 during testing. For comparison with VPM, we implement a baseline which learns global feature through the combination of cross-entropy loss and triplet loss. We also implement a part-based feature learning method PCB \cite{PCB_ECCV}. For fair comparison, we enhance PCB with an extra triplet loss during training, and achieve slightly higher performance than \cite{PCB_ECCV}. The results are summarized in Table \ref{tab:large}.

\textbf{VPM significantly improves partial re-ID performance over the baseline.} On Market-1501, VPM surpasses the baseline by +6.4\%, +5.4\%, +4.3\%, +4.4\%, +4.6\% +6.2\% rank-1 accuracy and +4.4\%, +4.6\%, +8.0\%, + 8.6\%, +10.9\%, +13.1\% mAP when $\gamma$ is set from 0.5 to 1, respectively. 
The superiority of VPM against  the baseline, which learns a global feature representation, is derived from two-fold benefit. On the one hand, VPM learns region-level features and benefits from fine-grained information. On the other hand, with visibility awareness, VPM conducts a region-level alignment and eliminates the distracting noises originated from  unshared regions. 

\textbf{VPM increases the robustness of part features under partial re-ID scenario.} Comparing VPM with PCB, a state-of-the-art part feature learning method for holistic person re-ID task, we observe that as $\gamma$ decreases, the retrieval accuracy achieved by PCB dramatically drops (\emph{e.g.}, 0.9\% rank-1 accuracy at $\gamma=0.5$ ), implying that PCB is extremely vulnerable to the spatial misalignment in partial re-ID.  By contrast, the retrieval accuracy achieved by VPM decreases much slower as $\gamma$ decreases. We infer that VPM facilitates region-to-region comparison within shared regions of two images and thus gains strong robustness.

We also notice that under $\gamma = 1.0$, \emph{i.e.}, the holistic person re-ID scenario, VPM achieves comparable retrieval accuracy with PCB.

\begin{figure}[t]
\setlength{\abovecaptionskip}{-1.2cm} 
\setlength{\belowcaptionskip}{-0.4cm}
\begin{center}
\includegraphics[width=1.0\linewidth]{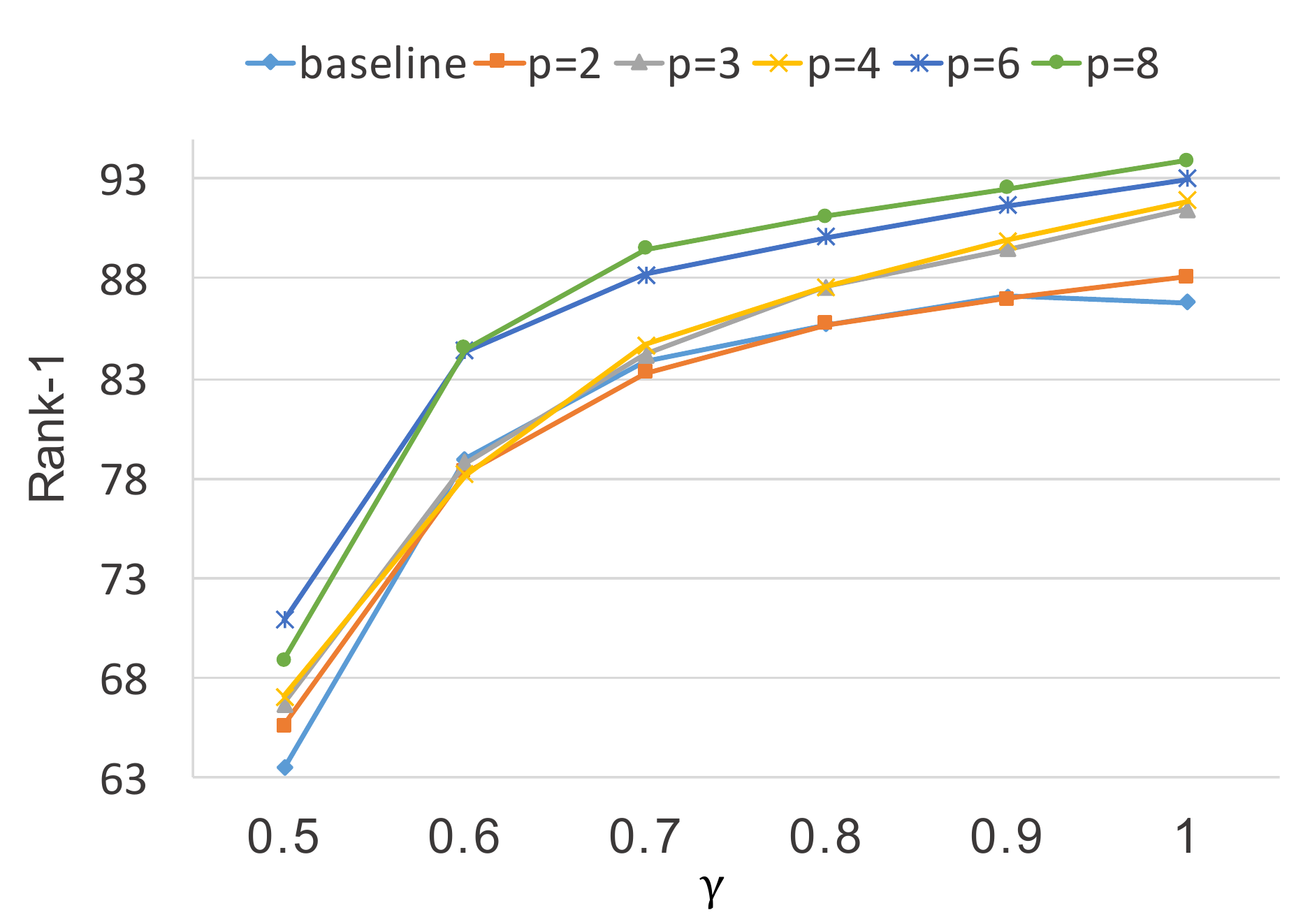}
\end{center}
   \caption{Impact of $p$ on the partial-reID accuracy. We set $p$ to 2,3,4,6 and 8, respectively. We use Market-1501 for training and differ the crop ratio $\gamma$ during testing.}
\label{fig:numbers}
\end{figure}
In Table \ref{tab:large}, we use 6 pre-defined parts to construct VPM. Moreover, we analyze the impact of the part numbers $p$ on Market-1501, with results shown in Fig. \ref{fig:numbers}. Under all settings of $p$ and $\gamma$, VPM consistently surpasses the baseline, which further confirms the superiority of VPM. We also observe that larger $p$ generally brings higher (rank-1) retrieval accuracy. Larger $p$ allows VPM to learn the region-level features in finer granularity and thus benefits the discriminative ability, which is consistent with the observation in holistic person re-ID works \cite{Zhao2017Deeply, PCB_ECCV}. Larger $p$ also allows more accurate region alignment when comparing a partial person image against a holistic one. We suggest choosing $p$ with the joint consideration of retrieval accuracy and computing efficiency, and set $p=6$ in most of our experiments (if not specially mentioned). 

\subsection{Comparison with the state of the art}\label{sec:sota}
We compare VPM with the state-of-the-art methods on two public datasets, \emph{i.e.}, Partial-REID and Partial-iLIDS. We train three different versions of VPM with different crop strategies for preparing training patches, \emph{i.e.}, top crop (the top regions are always visible), bottom crop (the bottom regions are always visible) and bilateral crop (top crop + bottom crop).  The results are presented in Table \ref{tab:sota}, from which two observations are drawn.

First, comparing three editions of VPM against each other, we find that the crop strategy matters to VPM. On Partial-iLIDS, all query images of which are cropped from the top side of holistic pedestrian images, VPM (Top) achieves the highest retrieval accuracy. On Partial-REID, which contains images cropped from different directions, VPM (Bilateral) achieves the highest retrieval accuracy.  VPM (Bottom) always performs the worst due to two reasons. First, retaining the bottom regions severely deviates from the testing condition. Second, the bottom regions (mainly containing legs) inherently offers relatively weak discriminative clues. We note that when solving the problem of partial-reID, the realistic partial condition is usually estimable. We recommend analyzing the partial condition and choosing a similar crop strategy for training VPM. That being said, VPM is general in that it is able to cooperate with various crop strategies. 

\setlength{\tabcolsep}{4.2pt}
\begin{table}[t]
\begin{center}
\begin{tabular}{l|cc|cc}
\hline
\multicolumn{1}{l|}{\multirow{2}{*}{Methods}}&\multicolumn{2}{c|}{Partial-REID}&\multicolumn{2}{c}{Partial-iLIDS}\\
\cline{2-5}
\multicolumn{1}{c|}{}&R-1&R-3&R-1&R-3\\
\hline
MTRC \cite{DBLP:journals/pami/LiaoJL13}                   &23.7 &27.3  &17.7&26.1\\
AMC+SWM \cite{DBLP:conf/iccv/ZhengLXLLG15}            &37.3 &46.0   &21.0&32.8\\
DSR \cite{HeLX2018Partial}                       &50.7 &70.0  &58.8  &67.2\\
SFR \cite{DBLP:journals/corr/abs-1810-07399}                        &56.9 & 78.5  &63.9 &74.8\\
\hline
VPM (Bottom)          &53.2 & 73.2  & 53.6  &  62.3\\      
VPM (Top)             &64.3 & \textbf{83.6}  &\textbf{67.2}  & \textbf{76.5}\\
VPM (Bilateral)      &\textbf{67.7}  &{81.9}  &65.5 &74.8  \\

\hline
\end{tabular}
\caption{Evaluation of VPM on Partial-REID and Partial-iLIDS. Three VPMs trained with different crop strategies are evaluated.}
\label{tab:sota}
\end{center}
\setlength{\abovecaptionskip}{0cm} 
\setlength{\belowcaptionskip}{0pt} 
\end{table}
Second, given appropriate crop strategies, VPM achieves very competitive performance compared with the state of the art. On Partial-REID, VPM (Bilateral) surpasses the strongest competing method SFR by +10.6\% Rank-1 accuracy. On Partial-iLIDS, VPM (Top) surpasses SFR by +3.3\% Rank-1 accuracy. Even with no prior knowledge of partial condition on testing set, we may eclectically choose VPM (Bilateral), which considers both top and down occlusions and thus maintains stronger robustness. 

\subsection{The importance of self-supervision}\label{sec:importance}
We conduct ablation study to analyze the impact of self-supervision on VPM. 
We train four \emph{Malfunctioned} VPM for comparison:
\begin{itemize}
\item MVPM-1 is trained as a normal VPM, but abandons the visibility awareness during testing, \emph{i.e.}, MVPM-1 concludes the overall distance with all region-level features, even if some regions are invisible. 
\item MVPM-2 abandons self-supervision on triplet loss during training, \emph{i.e.}, all region features equally contribute to deducing the triplet loss $L_{tri}$. 
\item MVPM-3 abandons self-supervision on identification loss $L_{ID}$ during training, \emph{i.e.}, all region features are supervised by the training identity label through $L_{ID}$.
\item MVPM-4 abandons self-supervision on both triplet loss and identification loss.
\end{itemize}

Moreover, we also analyze the impact of  two types of losses (cross-entropy loss and triplet loss) in training VPM. The results are summarized in Table \ref{tab:ablation}, from which we draw three observations.

\setlength{\tabcolsep}{4.2pt}
\begin{table}[t]
\begin{center}
\begin{tabular}{l|ccc|ccc}
\hline
\multicolumn{1}{l|}{\multirow{2}{*}{Methods}}&\multicolumn{3}{c|}{Partial-iLIDS}&\multicolumn{3}{c}{Market-1501}\\
\cline{2-7}
\multicolumn{1}{c|}{}&R-1&R-3&R-5&R-1&R-5&mAP\\
\hline

VPM.             &67.2  & 76.5& 82.4 & 93.0&97.8&80.8\\
VPM (no triplet) & 57.1 &73.9 &79.0 &91.3&97.0&77.8\\
\hline
MVPM-1    &63.0&74.8&82.4        &   93.0 & 96.3&79.7\\  
MVPM-2     &61.3 &73.1 &79.0      &  92.8&97.4&80.1\\  
MVPM-3     &58.8 & 74.8 & 82.4    &   91.4&96.5&75.5\\
MVPM-4     &59.7 &74.8 & 78.2     &    90.4&96.6&75.7\\

\hline
\end{tabular}
\caption{Ablation study on VPM. ``VPM (no triplet)'' is trained with no triplet loss. On Market-1501, we only analyze the holistic person re-ID mode.}
\label{tab:ablation}
\end{center}
\setlength{\abovecaptionskip}{0cm} 
\setlength{\belowcaptionskip}{0pt} 
\end{table}

First, appending an extra triplet loss enhances the discriminative ability of the learned features, under both partial re-ID scenario (Partial-iLIDS) and holistic person re-ID scenario (Market-1501). This observation is consistent with prior works \cite{Hermans2017DefenseTriplet, zhong2018generalizing, zhang2017alignedreid, YuDBZXB18} on holistic person re-ID task.

Second, comparing ``MVPM-1'' with ``VPM'', we observe a dramatic performance decrease on Partial-iLIDS. Both models are trained in exactly the same procedure. The difference is that ``MPVM-1'' employs all the region features to conclude the overall distance, while VPM focuses on the shared regions between two images. On Market-1501, all the regions are visible and two models achieves very close retrieval accuracy. We thus infer that the visibility awareness is critical for VPM under partial re-ID scenario. 

Third, comparing last three editions of MVPM with ``VPM'' as well as ``MVPM-1'', we observe further performance decreases on Partial-iLIDS.  The last three editions abandon self-supervision to regularize the learning of region-level features (either on the cross-entropy loss or triplet loss or both). Learning features from invisible regions brings about larger sample noises. Consequentially, the learned region features are significantly compromised. We thus conclude that enforcing VPM to focus on visible regions through self-supervision is critical for learning region features.

\subsection{Visualization of discovered regions}
\begin{figure}[t]
\setlength{\abovecaptionskip}{-0.2cm} 
\setlength{\belowcaptionskip}{-0.5cm}
\begin{center}
\includegraphics[width=0.8\linewidth]{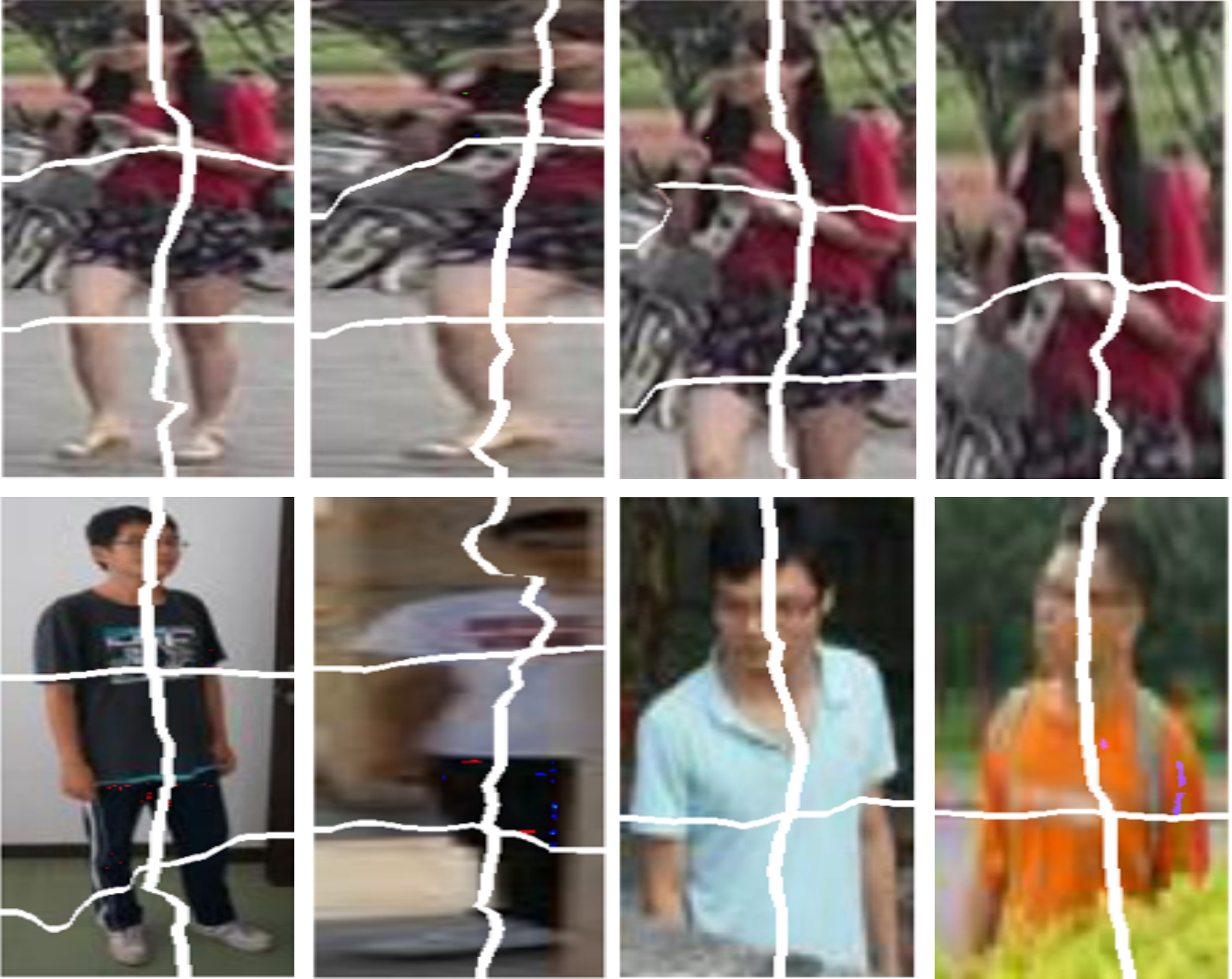}
\end{center}
   \caption{Region visualization. We train VPM with $3\times2$ pre-defined regions. For each image, VPM discovers 6 regions with 6 probability maps, as detailed in Section \ref{sec:structure}. For better visualization, we assign each pixel to its closest region and achieve the partitioning effect. Images on the first and the second row are from (synthetic) Market-1501 and Partial-REID, respectively.}
\label{fig:visual}
\end{figure}
We visualize the regions discovered by VPM (the region locator, in particular) in Fig. \ref{fig:visual}. We use $p=3\times 2$ pre-defined regions to facilitate both horizontal and vertical visibility awareness. 
It is observed that VPM conducts adaptive partition with visibility awareness. Given holistic images (the first column), VPM successfully discovers all the $3\times2$ regions. Given partial pedestrian images with horizontal occlusion (the second column), VPM favors the dominating regions (left regions in Fig. \ref{fig:visual}). Given partial pedestrian images with lower-body occluded (the last two columns), VPM roughly discovers 4 visible regions, and perceives that the bottom 2 regions are invisible. These observations confirm that VPM gains robust region-level visibility awareness and is capable to locate the visible regions through self-supervised learning.

\section{Conclusion}
In this paper, we propose a region-based feature learning method, VPM for partial re-ID task. Given a set of pre-defined regions on the holistic pedestrian image, VPM learns to perceive which regions are visible on a partial image through self-supervision. VPM locates each region on the convolutional feature maps and then extracts region-level features. With visibility awareness, VPM compares two pedestrian images with focus on their shared regions and correspondingly suppresses the severe spatial misalignment in partial re-ID. Experimental results confirm that VPM surpasses both the global feature learning baseline and part-based convolutional methods, and the achieved performance is on par with the state of the art.  

{\small
\bibliographystyle{ieee}
\bibliography{egbib}
}

\end{document}